# Artificial Geographically Weighted Neural Network: A Novel Framework for Spatial Analysis with Geographically Weighted Layers

Jianfei Cao[a], Dongchao Wang[a,*]

[a] College of Geography and Environment, Shandong Normal University, Jinan 250358, China; caojianfei@sdnu.edu.cn (J.C.); 2022010116@stu.sdnu.edu.cn (D.W.).

[*] Corresponding author: (D.W.) 2022010116@stu.sdnu.edu.cn.

**Abstract**

Geographically Weighted Regression (GWR) is a widely recognized technique for modeling spatial heterogeneity. However, it is commonly assumed that the relationships between dependent and independent variables are linear. To overcome this limitation, we propose an Artificial Geographically Weighted Neural Network (AGWNN), a novel framework that integrates geographically weighted techniques with neural networks to capture complex nonlinear spatial relationships. Central to this framework is the Geographically Weighted Layer (GWL), a specialized component designed to encode spatial heterogeneity within the neural network architecture. To rigorously evaluate the performance of AGWNN, we conducted comprehensive experiments using both simulated datasets and real-world case studies. Our results demonstrate that AGWNN significantly outperforms traditional GWR and standard Artificial Neural Networks (ANNs) in terms of model fitting accuracy. Notably, AGWNN excels in modeling intricate nonlinear relationships and effectively identifies complex spatial heterogeneity patterns, offering a robust and versatile tool for advanced spatial analysis.

## 1. Introduction

Spatial heterogeneity refers to the alteration in the relationship or structure between variables due to variations in geographical location (Anselin 1989, Goodchild 2004). The consideration of spatial heterogeneity is essential for the construction of spatial regression models, as it facilitates a more comprehensive understanding as well as analysis of the intricate and diverse nature of spatial data (Brunsdon *et al.* 1996, Fotheringham *et al.* 1998). The computation of spatial heterogeneity has become a current research hotspot in fields such as geography and geographical information science (Fotheringham *et al.* 2002, Beale *et al.* 2010).

GWR model is a specialized regression framework developed specifically to address spatial heterogeneity (Brunsdon *et al.* 1996, Fotheringham *et al.* 2002) and has been widely applied in various fields. Based on the traditional regression frameworks, GWR introduces the spatial weight matrix, enabling the model parameters to change in accordance with the variations of spatial locations. GWR has gained widespread recognition and application among scholars in various domains, including air pollution analysis (Zhou *et al.* 2019), geochemical research (Nazarpour *et al.* 2022), climate monitoring (Wei *et al.* 2023), real estate valuation (Dambon *et al.* 2021), and ecological risk assessment (Liu *et al.* 2019). Many scholars have been continuously improving the theory and methods of the GWR model. Brunsdon et al. (1999) introduced a hybrid GWR model that simultaneously incorporates both global and local information by integrating global and local parameters within the framework. Subsequently, Fotheringham *et al.* (2017) put forward a Multiscale Geographically Weighted Regression (MGWR) approach to estimate the regression parameters at diverse spatial scales, thereby accurately presenting the heterogeneous characteristics. Huang *et al.* (2010) developed the Geographically and Temporally Weighted Regression (GTWR) by calculating the spatial and temporal weight separately, based on the assumption that the time dimension and spatial dimension are independent of each other. Additionally, a crucial element of the GWR model is the spatial weight matrix, which illustrates local spatial relationships. To improve the expression of local spatial relationships in GWR, Lu *et al.* (2014, 2016) endeavored to calculate the heterogeneity characteristics of spatial relationships with irregular distributions in the geospatial domain using non-Euclidean distance metrics.

Despite these advances, GWR and its derivatives remain constrained by their reliance on linear partial least squares assumptions, limiting their capacity to model pervasive nonlinear relationships in geospatial data(Wheeler and Tiefelsdorf 2005, Hagenauer and Helbich 2022). In recent years, scholars have begun to investigate methods that combine geographical weighting with Neural Network (NN). The Geographically Neural Network Weighted Regression (GNNWR, Du *et al.* 2020) and Geographically and Temporally Neural Network Weighted Regression (GTNNWR, Wu *et al.* 2021) that combine Ordinary Least Squares (OLS) and NN are proposed to estimate spatial heterogeneity based on the concept similar to GWR, which have been successfully applied in fields such as marine ecological modeling, spatial estimation of atmospheric pollutants, and urban housing prices research (Du *et al.* 2021, Wang *et al.* 2022, Liang *et al.* 2023). However, it is important to note that the models primarily focus on the nonlinear expression of spatial weights, rather than the nonlinear relationship between variables. Hagenauer and Helbich (2022) constructed a Geographically Weighted Artificial Neural Network (GWANN), which assigned the output neurons of Artificial Neural Network (ANN) to corresponding locations in geographical space, to capture spatial heterogeneity relationships. However, the network structure fails to consider spatially varying coefficients, resulting in limited explanatory power compared to GWR.

To overcome these limitations, this study proposes an Artificial Geographically Weighted Neural Network (AGWNN) to capture spatial heterogeneity information, which is constructed with input layer, hidden layers, Geographically Weighted Layer (GWL), and output layer. The organization of this paper is structured as follows: Section 2 briefly presents the associated models and introduces the innovative geographical neural network model proposed in this paper; In Section 3, various models are compared and demonstrated using simulated and real datasets; The results and further discussions based on this research, along with recommendations, are provided in Section 4. Finally, this paper summarizes the findings in Section 5.

## 2. Methodology

### 2.1 ANN model

ANN can arbitrarily approximate any nonlinear function (Papapicco *et al.* 2022). In a typical ANN, the network structure comprises an input layer, hidden layers, and an output layer. Forward propagation begins at the input layer, and through the network's connection weights and activation functions, the input signal is propagated layer-by-layer until it reaches the output layer, which results in the network's predicted output.

The role of the activation function is to transform the original relationship between the input and output, thereby enhancing the ANN's ability to classify or fit complex relationships. The input signal is transformed into an output signal within the neuron, and the specific calculation formula is as follows:

$$net_j = \sum_{i=1}^{n} w_i input_i + b_j \qquad (1)$$

$$output_j = f(net_j) \qquad (2)$$

where *w* represents the weight coefficients, *b* represents the bias coefficients, and *f* denotes the activation function.

The error function compares the predicted output with the expected output, calculating the difference between them. Commonly used error functions include Residual Sum of Squares (RSS), Mean Squared Error (MSE), and Cross-Entropy Error. When employing an ANN to handle regression problems, the model's loss function typically uses the RSS. Dividing by 2 in the formula make it easier to calculate the derivative. The specific formula is as follows:

$$Loss = \frac{1}{2}\sum_{i=1}^{n}(target_i - output_i)^2 \qquad (3)$$

Where *Loss* denotes the amount of loss, *target* denotes the amount of goal, and *output* denotes the amount of output.

**2.2 GWR model**

To incorporate geographical factors into regression models, scholars proposed spatially varying coefficient regression models, and the GWR model was further established. It incorporates geographical location features to express spatial variations in regression coefficients. The formula is as follows:

$$y_i = \beta_0(u_i, v_i) + \sum_{m=1}^{p} \beta_m(u_i, v_i) x_{im} + \varepsilon_i, \quad i = 1,2,\cdots,n \qquad (4)$$

In the above equation, the positional coordinates (typically represented using latitude and longitude) at the *ith* sample point are denoted as $(u_i, v_i)$. The *mth* regression coefficient at the *ith* sample point is represented as $\beta_m(u_i, v_i)$. The regression coefficients can be expressed as a function of location, where the independent variables are $u_i$ and $v_i$.

$$y_i = X_i \beta_i + \varepsilon_i \qquad (5)$$

The estimation of the regression coefficients can be used to estimate the regression value $\hat{y}_i$ at sample point *i*, denoted as follows:

$$\hat{y}_i = X_i \hat{\beta}_i = X_i (X^T W_i X)^{-1} X^T W_i Y \qquad (6)$$

In GWR, the spatial weights $W_i$ serve as a tool to measure geographical proximity. In this article, we introduce the hat matrix specific to GWR, denoted by the formula for the hat matrix $S$ as follows:

$$S_i = X_i (X^T W_i X)^{-1} X^T W_i \qquad (7)$$

The unbiased estimate of the variance of the error term is denoted as $\hat{\sigma}^2$.

$$RSS = \sum_{i=1}^{n} \varepsilon_i^2 = e^T e = Y^T (I - S)^T (I - S) Y \qquad (8)$$

$$\hat{\sigma}^2 = \frac{RSS}{n-2tr(S)+tr(S^TS)} \approx \frac{RSS}{n-tr(S)} \tag{9}$$

The spatial bandwidth in GWR is classified into two types: fixed and adaptive. The spatial weight kernel functions typically include the Gaussian function and Gaussian-like functions such as the bi-square function. The selection of the optimal spatial bandwidth directly impacts the estimation accuracy of the GWR model. Different diagnostic criteria yield varying optimal bandwidths. In the field of GWR, the Akaike Information Criterion corrected (AICc) criterion is commonly used for selecting the optimal bandwidth (Brunsdon *et al.* 2002). The mathematical formula for *AICc* is as follows:

$$AICc = 2n \ln(\hat{\sigma}^2) + n \ln(2\pi) + n \left[\frac{n+tr(S)}{n-2-tr(S)}\right] \tag{10}$$

It incorporates geographically weighting in regression analysis by considering spatial heterogeneity and spatial correlation. GWR is well-suited for addressing issues related to spatial heterogeneity, effectively exploring spatial patterns of variation in geographical data.

**2.3 Artificial geographically weighted neural network**

The proposed AGWNN (see Figure 1) integrates the functionality of ANN and GWR, aiming to explore a method that can address both nonlinearity and spatial heterogeneity.

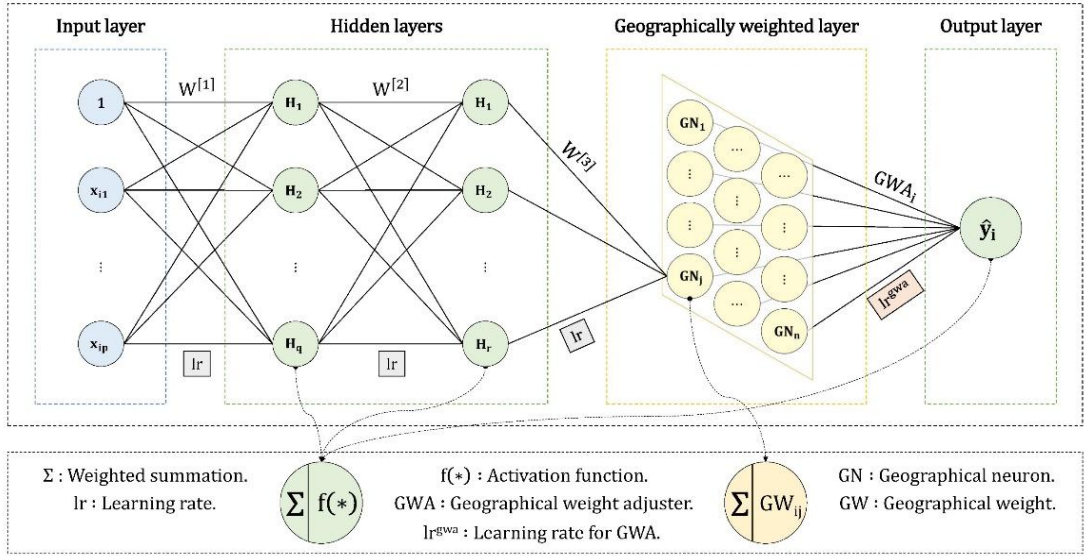

Figure 1. The architecture of AGWNN model.

**2.3.1 Geographically weighted layer**

The existing layers in neural networks primarily focus on transforming input signals globally, without considering the influence of spatial positions on the output signal. To address this, we have designed a specialized GWL that utilizes geographical weights for measuring spatial correlation.

Empirical spatial models explain this relationship using a typically "bell-shaped" function, often represented by a Gaussian kernel function, as shown in the formula below:

$$gw_{ij} = e^{-(dist_{ij}/bw)^2} \tag{11}$$

In the above equation, *gw*$_{ij}$ represents the spatial weight between samples *i* and *j*, *dist*$_{ij}$ is their spatial distance, and *bw* is the spatial bandwidth.

Each neuron of GWL is assigned to a location in geographical space, and these neurons called Geographical Neurons (GN, see Figure 1). The specific formula for GWL is as follows:

$$GWL: \quad f(net_{ij}) = net_{ij} \times gw_{ij} \tag{12}$$

where *net*$_{ij}$ denotes the node value of the *ith* input sample on the *jth* neuron of GWL.

The GWL alter the distribution characteristics of the signals, compressing the activation trend line in the direction away from the target sample, resulting in a Gaussian distribution in space. For simulating complex spatial weights, we have designed the Geographical Weight Adjuster (GWA), which is a specialized tuning mechanism. The adjustment range is constrained between 0 and 1 (Lu *et al.* 2017), with the adjustment magnitude primarily determined by the learning rate of GWA. The specific formula is as follows:

$$\begin{cases} agw_{ij} = gw_{ij} \times gwa_{ij} \\ gwa_{ij} \in (0,1) \\ \Delta gwa_{ij} = f(\nabla gwa_{ij}, lr^{gwa}) \end{cases} \tag{13}$$

In the above equation, *agw*$_{ij}$ denotes the corrected spatial weight of the *ith* input sample on the *jth* geographical neuron, *gwa*$_{ij}$ represents its GWA, $\nabla gwa_{ij}$ and $\Delta gwa_{ij}$ are the GWA before and after learning respectively, *lr*$^{gwa}$ is the learning rate for GWA, and *f* is the is the gradient descent learning algorithm.

**2.3.2 AGWNN framework and principle**

The proposed AGWNN model comprises four modules: the input layer, the hidden layers, the geographically weighted layer, and the output layer. The input layer is assigned to transmit the sample independent variables. The hidden layers are employed to learn nonlinear relationships. The GWL is responsible for the spatial heterogeneity calculation. The output layer integrates the signals to estimate the dependent variable.

Here, we predefine the meanings of symbols and variables for later use. *p* represents the number of independent variables. *q* and *r* denote the numbers of neurons in the ordinary hidden layers. The superscript *[li]* indicates the network layer index of the element. The subscript *i* represents the sequence variable of the training set. The subscript *j* represents the sequence variable of the neurons in the layer.

To facilitate matrix operations, modifications are required on the original matrices. The first column of the independent variable matrix $X_{origin}$ is augmented with a unit column vector $I$.

$$X = [I \quad X_{origin}] = \begin{bmatrix} X_1 \\ \vdots \\ X_n \end{bmatrix} = \begin{bmatrix} 1 & x_{11} & \cdots & x_{1p} \\ \vdots & \vdots & \ddots & \vdots \\ 1 & x_{n1} & \cdots & x_{np} \end{bmatrix} \tag{14}$$

The spatial weight matrix for all samples in the training set is represented as $GW$.

$$GW = \begin{bmatrix} GW_1 \\ \vdots \\ GW_n \end{bmatrix} = \begin{bmatrix} gw_{11} & \cdots & gw_{1n} \\ \vdots & \ddots & \vdots \\ gw_{n1} & \cdots & gw_{nn} \end{bmatrix} \tag{15}$$

The spatial weight for the *ith* training sample at the *jth* position in the GWL is denoted as $gw_{ij}$. The spatial weight matrix for all neurons in the layer can be represented as $GW_i$.

The processing of input signals in the AGWNN method can be represented by the following formula:

$$\hat{y}_i = f^{[5]}\left(\left(\left(f^{[3]}(f^{[2]}(X_i W^{[1]})W^{[2]})W^{[3]}\right) \times GW_i^T\right) GWA_i\right) \quad (16)$$

where $W$ represent the ordinary weights, and $f(*)$ denote the activation functions.

After letting $f(*) = *$, AGWNN metamorphoses into an ordinary local spatial regression model, which is able to visualize the local regression coefficients. The signal processing process is as follows:

$$\hat{y}_i = \left(\left(X_i W^{[1]} W^{[2]} W^{[3]}\right) \times GW_i^T\right) GWA_i = X_i \left(W^{[1]} W^{[2]} W^{[3]}\right)(GW_i \times GWA_i) \quad (17)$$

It is evident that the spatial regression coefficient equation of the AGWNN is as follows:

$$\hat{\beta}_i = \left(W^{[1]} W^{[2]} W^{[3]}\right)(GW_i \times GWA_i) \quad (18)$$

Benefiting from the two modes, the AGWNN can not only detect spatial nonlinear relationships but also visualize local regression coefficients.

**2.3.3 Model optimization and evaluation**

The AGWNN model inherently incorporates numerous hyperparameters necessary for neural network learning, and including the spatial bandwidth among them would increase the computational cost of model training. Given that the AGWNN inherits all the characteristics of GWR, leveraging GWR for bandwidth selection can save significant computational resources and time.

The output layer of AGWNN consists of a single neuron, and the error function utilizes the classical L2-norm loss function (Equation (3)). In the field of gradient descent, numerous new algorithms have emerged for performing iterative updates. In practical applications, algorithmic paradigms are not strictly defined, and multiple algorithms can be used simultaneously. The Mini-Batch Stochastic Gradient Descent (SGD) is widely employed in the training set allocation stage (Li *et al.* 2014). The NAdam algorithm (Dozat 2016) applies stronger constraints on the learning rate and has a more direct impact on gradient updates.

The learning rate controls the iteration step size of hyperparameters in a neural network. An excessively high value can lead to non-convergence of the learning process, while a too low value can increase the learning cost. In the AGWNN model, the learning rate *lr* is used to tune the global learning rate of the neural network. On the other hand, the learning rate *lr$^{gwa}$* between the GWL and the output layer is specifically responsible for fine-tuning the geographical weights. To prevent excessive adjustments of geographical weights, *lr$^{gwa}$* is typically set to a value several orders of magnitude lower than *lr*. This value is determined based on the specific data context.

Utilizing the machine learning approach to train the AGWNN model, it is crucial to select appropriate evaluation metrics for regression problems. Common evaluation metrics include MAE, Mean Squared Error (MSE), Root Mean Squared Error (RMSE), R-Squared ($R^2$), and adjusted R-Squared (adj$R^2$). Considering the GWR-like characteristics of AGWNN, the AICc is suitable as an evaluation metric, particularly suitable for assessing results in spatially heterogeneous datasets.

## 3. Experiments

This study validates the superiority of the AGWNN method in handling nonlinearity and spatial heterogeneity using both simulated and real datasets. GWR utilizes a fixed Gaussian spatial kernel function to assist to construct the GW matrix, with bandwidth optimization which is based on minimizing AICc.

The ANN employs a three-layer network structure, and its activation function "softsign" is utilized for both the hidden and output layers. Weights and thresholds initialization follow the Xavier initialization method (Glorot and Bengio 2010), and the NAdam algorithm is selected as the gradient descent optimizer. The dataset allocation is achieved through K-fold cross-validation (Islam *et al.* 2021). Model selection includes an early-stopping strategy (Lee *et al.* 2021) to prevent overfitting and mitigate the risk of getting stuck into local optimal solutions.

### 3.1 Modeling of simulated dataset

The experiment aims to compare the performance of Multiple Linear Regression (MLR), ANN, GWR, GWANN, GNNWR, and AGWNN on six synthetic datasets. This study has three primary objectives: to demonstrate the feasibility of AGWNN to simultaneously detect nonlinearity and spatial heterogeneity; to verify AGWNN's superior capability for rebuilding spatial models; and to explore the AGWNN's predictive robustness.

### 3.1.2 Synthetic dataset

The coefficients are designed to represent different levels of spatial heterogeneity. They are generated as follows:

$$\beta_a = 2 + \frac{\varepsilon_i}{4} \tag{19}$$

$$\beta_b(u_i, v_i) = \frac{1}{10}(u_i + v_i) + \frac{\varepsilon_i}{4} \tag{20}$$

$$\beta_c(u_i, v_i) = 2\left(1 + sin\left(\frac{u_i\pi}{10}\right)^2 - cos\left(\frac{v_i\pi}{10}\right)^2\right) + \frac{\varepsilon_i}{4} \tag{21}$$

$$\beta_d(u_i, v_i) = 2\left|cos\frac{u_i\pi}{10} + cos\frac{v_i\pi}{10}\right| + \frac{\varepsilon_i}{4} \tag{22}$$

where $(u_i, v_i)$ represents the spatial coordinates of sample point *i*. $\varepsilon_i$ is the error term following a normal distribution $N(0,0.25)$. $\beta_a$ denotes a constant surface with no spatial variation. $\beta_b$ represents a linear trend surface. $\beta_c$ and $\beta_d$ represent nonlinear trend surfaces. As shown clearly in Figure 2, the spatial heterogeneity of $\beta_d$ (Var=1.458) is the highest, followed by $\beta_c$ (Var=1.053), while $\beta_b$ (Var=0.756) exhibits moderate spatial variation, and $\beta_a$ (Var=0.004) shows no spatial heterogeneity at all.

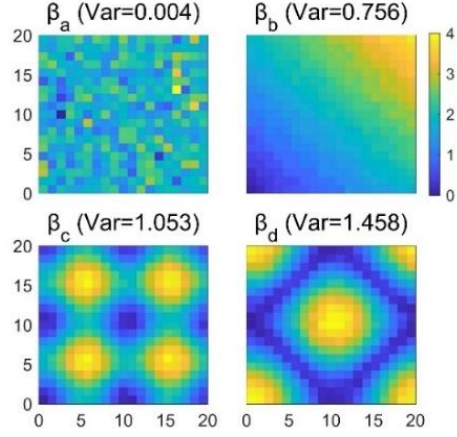

Figure 2. Coefficients' surfaces with different levels of spatial heterogeneity. (The variances of $\beta$ in parentheses indicate the status of spatial heterogeneity)

In this experiment, different coefficients are combined to create four synthetic datasets. The formulas are as follows:

$$y_i^1 = \beta_a + \beta_b(u_i, v_i)x_{i1} + \beta_c(u_i, v_i)x_{i2} \tag{23}$$

$$y_i^2 = \beta_a + \beta_b(u_i, v_i)x_{i1} + \beta_d(u_i, v_i)x_{i2} \tag{24}$$

$$y_i^3 = \beta_a + tanh(\beta_b(u_i, v_i)x_{i1}) + tanh(\beta_c(u_i, v_i)x_{i2}) \tag{25}$$

$$y_i^4 = \beta_a + tanh(\beta_b(u_i, v_i)x_{i1}) + tanh(\beta_d(u_i, v_i)x_{i2}) \tag{26}$$

For all functions, $(u_i, v_i)$ represents the location of grid cell *i*, and $x_i$ is a random variable following a uniform distribution $U(0,1)$. The coefficients $\beta_a$, $\beta_b$, $\beta_c$, and $\beta_d$ represent the respective coefficients for grid cell *i*. The first two functions simulate linear relationships between the dependent and independent variables, while the latter two functions use the hyperbolic tangent function to represent nonlinear relationships.

Based on the definitions of the coefficients, the first function (Equation (23)) represents a linear and low spatial heterogeneity scenario; the second function (Equation (24)) represents a linear and high spatial heterogeneity scenario; the third function (Equation (25)) represents a nonlinear and low spatial heterogeneity scenario; and the fourth function (Equation (26)) represents a nonlinear and high spatial heterogeneity scenario.

To intuitively demonstrate the AGWNN model's ability to rebuild spatial regression relationships, we combine all the coefficients to form a single function (Equation (27)) that incorporates various levels of spatial heterogeneity. This function simulates an original synthetic dataset, which is used for visual verification experiments of the local regression coefficients. Additionally, 1,000 random test datasets are simulated based on this function. The AGWNN model, trained using the original dataset, is then applied to these 1,000 test datasets to evaluate the predictive performance of the AGWNN model.

$$y_i^5 = \beta_a + \beta_b(u_i, v_i)x_{i1} + \beta_c(u_i, v_i)x_{i2} + \beta_d(u_i, v_i)x_{i3} \tag{27}$$

### 3.1.2 Diagnosis of AGWNN estimation capacity

The result of estimation errors (RMSE) for different models are showed in Figure 3. In the first and second rows, the global RMSE of ANN (0.769, 0.809) is slightly lower than that of MLR (0.769, 0.811), and the GWANN (0.147, 0.164) also similarly outperforms GWR (0.151, 0.172). However, as nonlinear features emerge, the performance rankings among the four models reverse. Relatively, the estimation performance (0.142, 0158, 0.164, 0.166) of the GNNWR model outperforms the previous four models. Due to the unique modeling capabilities in capturing nonlinear relationships and spatial heterogeneity, AGWNN exhibits the best performance for global and local estimation error among all models. When the relationships are highly nonlinear and the spatial variance of the coefficients is large, the RMSE differences (>0.029) between AGWNN and the other models become increasingly pronounced.

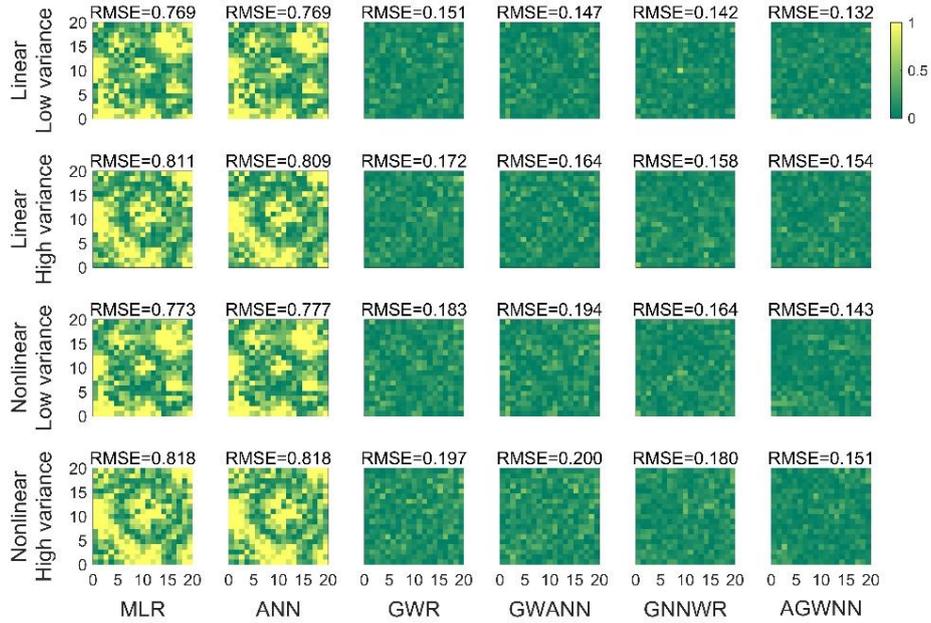

Figure 3. Comparison of estimation errors (RMSE) for different models. (The RMSEs on the top of the pseudo-color plot represents the global value, and the pseudo-color plot represents the local RMSE at different locations)

**3.1.3 Coefficients visualization capability for AGWNN**

Figure 4 provides a detailed comparison of the coefficient reconstruction ability among GWR, GNNWR, and AGWNN models. To eliminate the interference of the randomness of machine learning on the accuracy of the diagnostic results, the GNNWR and AGWNN models were trained 100 times and then averaged. The distribution of model regression coefficients is displayed using pseudo-color plots. GWR can only capture the general trends of the four beta values, resulting in fitting outcomes that exhibit significant noise (RMSE($\beta_a$)=1.013, RMSE($\beta_b$)=0.460, RMSE($\beta_c$)=0.497, RMSE($\beta_d$)=0.514); By contrast, GNNWR enhances the capability of coefficient visualization (RMSE($\beta_a$)=0.996, RMSE($\beta_b$)=0.452, RMSE($\beta_c$)=0.482, RMSE($\beta_d$)=0.509). On the other hand, AGWNN accurately reproduces the detailed variations in beta values, yielding the estimation results with minimal noise (RMSE($\beta_a$)=0.987, RMSE($\beta_b$)=0.447, RMSE($\beta_c$)=0.380, RMSE($\beta_d$)=0.507). Although all models exhibit less stable performance in estimating constant coefficient $\beta_a$ and linear coefficient $\beta_b$, AGWNN performs significantly better than GWR and GNNWR. It is particularly noteworthy that AGWNN precisely reproduces the four peaks of $\beta_c$ and the "diamond-shaped" trough of $\beta_d$. Therefore, AGWNN's coefficient reconstruction performance substantially exceeds that of other models and demonstrates excellent noise-smoothing capabilities.

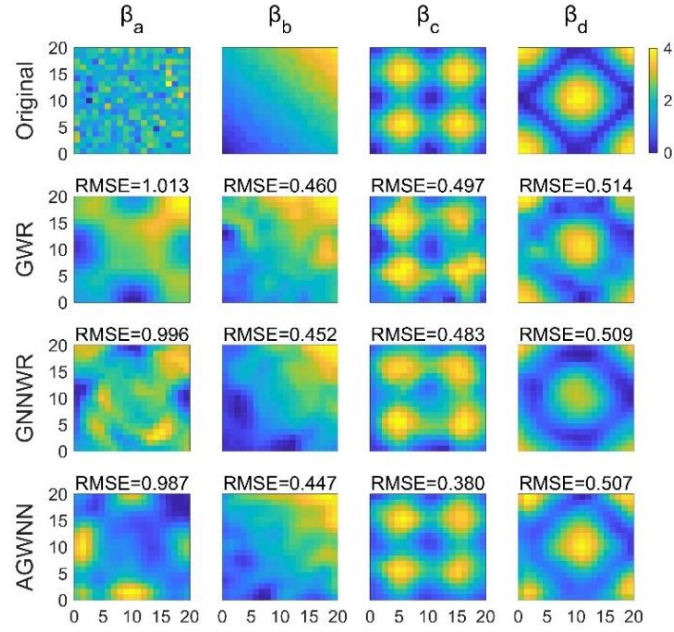

Figure 4. Comparison of the estimated coefficients ($\beta$) for different models. (The AGWNN and GNNWR models are trained 100 times and averaged to obtain results, the RMSEs on the top of the pseudo-color plot represents the global value, and the pseudo-color plot represents the local coefficients at different locations)

**3.1.4 Prediction performance of AGWNN**

In Table 1, we present the training metrics for the six models, including ENP, AICc, LOSS, RMSE, and $R^2$. The more complex the model structure, the greater the time cost. GNNWR is the longest (15.3), followed by AGWNN (13.4) and GWANN (6.4), and then GWR (3.9), ANN (0.3) and MLR (0.1). When comparing GWR and AGWNN, it is evident that AGWNN outperforms GWR by a significant margin. Notably, AICc decreases from 304.9 for GWR to 173.3 for AGWNN, a reduction of 131.6, indicating a substantial improvement in regression performance. AGWNN's $R^2$ is 0.989, representing an extremely high-precision fit to the data. Moreover, all metrics demonstrate outstanding performance upon upgrading from ANN to AGWNN, demonstrating the advanced parameter tuning capabilities within AGWNN. Overall, the models' performance can be categorized into three levels: the first level includes MLR (LOSS=181.0, RMSE=0.951, R2=0.489), ANN (LOSS=178.6, RMSE=0.945, R2=0.496) and GWR (LOSS=8.8, RMSE=0.210, R2=0.975); the second level includes GWANN (LOSS=8.0, RMSE=0.199, R2=0.978) and GNNWR (LOSS=5.8, RMSE=0.171, R2=0.984); and AGWNN (LOSS=3.8, RMSE=0.137, R2=0.989) is in the third level.

Conducting 1,000 prediction experiments with synthetic data, the results (Table 1) highlight AGWNN's exceptional predictive ability among the six models. The RMSE of the MLR and ANN models are basically the same (0.999), but the LOSS (199.8) and R2 (0.430) of the ANN are superior to the MLR's metrics (200.1, 0.429), and the predictive metrics of the GWR (47.2, 0.476, 0.864) are greatly improved. The prediction performance of the GWANN (44.4,0.459 ,0.872) is in the superior class, while GNNWR (43.2, 0.454, 0.876) shows a slight improvement. Comparatively, AGWNN has the best predictive performance among the six models, with LOSS, RMSE, and R2 of 39.2, 0.429, and 0.887, respectively.

Table 1. Training and predicting results for different models over 1,000 synthetic datasets.

| Model | Pattern | Time (second) | AICc | LOSS | RMSE | $R^2$ |
|---|---|---|---|---|---|---|
| MLR | Training | 0.1 | - | 181.0 | 0.951 | 0.489 |
| | Predicting | - | - | 200.1 | 0.999 | 0.429 |
| ANN | Training | 0.3 | - | 178.6 | 0.945 | 0.496 |
| | Predicting | - | - | 199.8 | 0.999 | 0.430 |
| GWR | Training | 3.9 | 304.9 | 8.8 | 0.210 | 0.975 |
| | Predicting | - | - | 47.2 | 0.476 | 0.864 |
| GWANN | Training | 6.4 | - | 8.0 | 0.199 | 0.978 |
| | Predicting | - | - | 44.4 | 0.459 | 0.872 |
| GNNWR | Training | 15.3 | 180.2 | 5.8 | 0.171 | 0.984 |
| | Predicting | - | - | 43.2 | 0.454 | 0.876 |
| AGWNN | Training | 13.4 | 173.3 | 3.8 | 0.137 | 0.989 |
| | Predicting | - | - | 39.2 | 0.429 | 0.887 |

Furthermore, the box plots (Figure 5) show the differences in statistical metrics such as RMSE and Pearson's r (PR) between the predicted and actual values. This further supports AGWNN's superiority over other models: MLR (RMSE=0.999, PR=0.697), ANN (RMSE=0.999, PR=0.698), GWR (RMSE=0.476, PR=0.967), GWANN (RMSE=0.459, PR=0.973) and GNNWR (RMSE=0.454, PR=0.975). AGWNN exhibits significantly better RMSE (0.429) and PR (0.978) than other models, further substantiating AGWNN's robustness in spatial modeling.

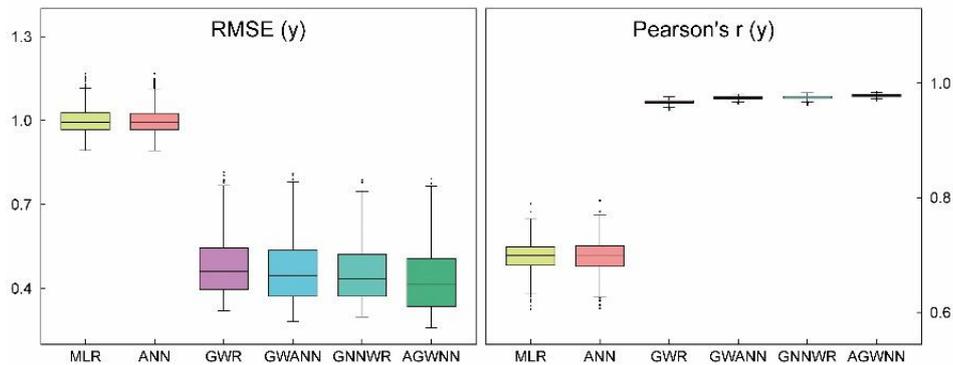

Figure 5. The distributions of RMSE and Pearson's r between predicted values and real values for different models over 1,000 synthetic datasets.

**3.1.5 Impact of training samples on AGWNN**

To understand the impact of the number of geographical neurons on AGWNN performance, we conducted a comparative experiment using a simulated dataset with 600 training epochs. We performed random subsampling on the observation samples and analyzed the changes in AGWNN's regression accuracy and efficiency for different training sample sizes. As shown in Figure 6, when the training sample size decreases, the prediction accuracy of AGWNN decreases. In contrast, its computational efficiency increases. This reverse relationship clearly indicates that the AGWNN model is highly sensitive to the training data size, and improving prediction accuracy requires sufficient sample support. However, an increase in the sample size also leads to a sharp decline in computational efficiency. Additionally, reducing the number of training samples weakens the resolution of the regression coefficients and increases the early-stopping epochs in neural network training. We need to find a balance between regression accuracy and computational efficiency to achieve optimal operation of the AGWNN model.

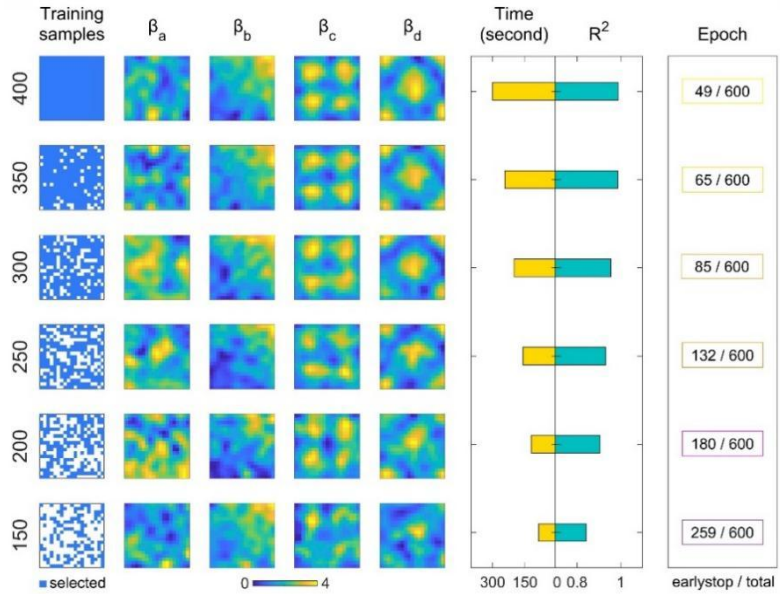

Figure 6. Variation of regression accuracy and efficiency of AGWNN with different number (400, 350, 300, 250, 200, 150) of training samples.

**3.2 Modeling of atmospheric environmental data**

Using real air quality data from the United States (U.S.) Environmental Protection Agency (EPA), we aim to validate the regression performance of AGWNN and assess its capability to detect spatial heterogeneity. AGWNN will be used to explore spatial heterogeneity among factors in the PM2.5 data modeling process and to analyze temporal patterns of PM2.5 concentration changes.

**3.2.2 Study region and dataset**

The empirical study (Figure 7) focuses on the Contiguous U.S. (CONUS), encompassing 48 states and the District of Columbia, excluding Alaska and Hawaii. Atmospheric pollution consists primarily of PM2.5 particles, which persist in the atmosphere and have wide-ranging and far-reaching impacts on human health and the environment (Jia *et al.* 2023). Remote sensing of the environment enables real-time, continuous monitoring of PM2.5 concentrations on a global scale (Christopher and Gupta 2020). The PM2.5 data from the U.S. EPA have been rigorously validated for data quality as a ground-based sensor monitoring product for the U.S. (Liu *et al.* 2005). This dataset provides important data support for researchers and policy makers to enhance understanding and addressing air pollution issues (Currie *et al.* 2023). We collected a total of 262,800 daily average PM2.5 records in 2019.

Estimating PM2.5 from remote sensing includes methods such as Aerosol Optical Depth (AOD) inversion (Hua *et al.* 2019, Xu *et al.* 2021) and atmospheric chemical transport modeling (Li *et al.* 2020). In this study, a typical strategy for estimating ground-level PM2.5 from satellite-derived AOD is used to establish a statistical relationship between AOD and PM2.5. The key parameters of the PM2.5 inversion model include AOD, the Normalized Difference Vegetation Index (NDVI), and meteorological conditions (e.g., temperature, humidity, wind speed). AOD provides global-scale atmospheric optical information (Gui *et al.* 2021). NDVI is used to assess the condition of surface vegetation and is correlated with PM2.5 concentrations (Jin *et al.* 2022). Meteorological conditions have a significant impact on PM2.5 concentrations (Zhang *et al.* 2020). In this experiment, we selected data for atmospheric environmental modeling from MODIS-AOD (https://lpdaac.usgs.gov/products/mcd19a2v061/), MODIS-NDVI (https://lpdaac.usgs.gov/products/myd13q1v061/), NOAA-RTMA (https://www.nco.ncep.noaa.gov/pmb/products/rtma/), and EPA-PM2.5 (https://aqs.epa.gov/aqsweb/airdata/download_files.html). Figure 7 depicts the spatial distribution pattern of the six variables in 2019.

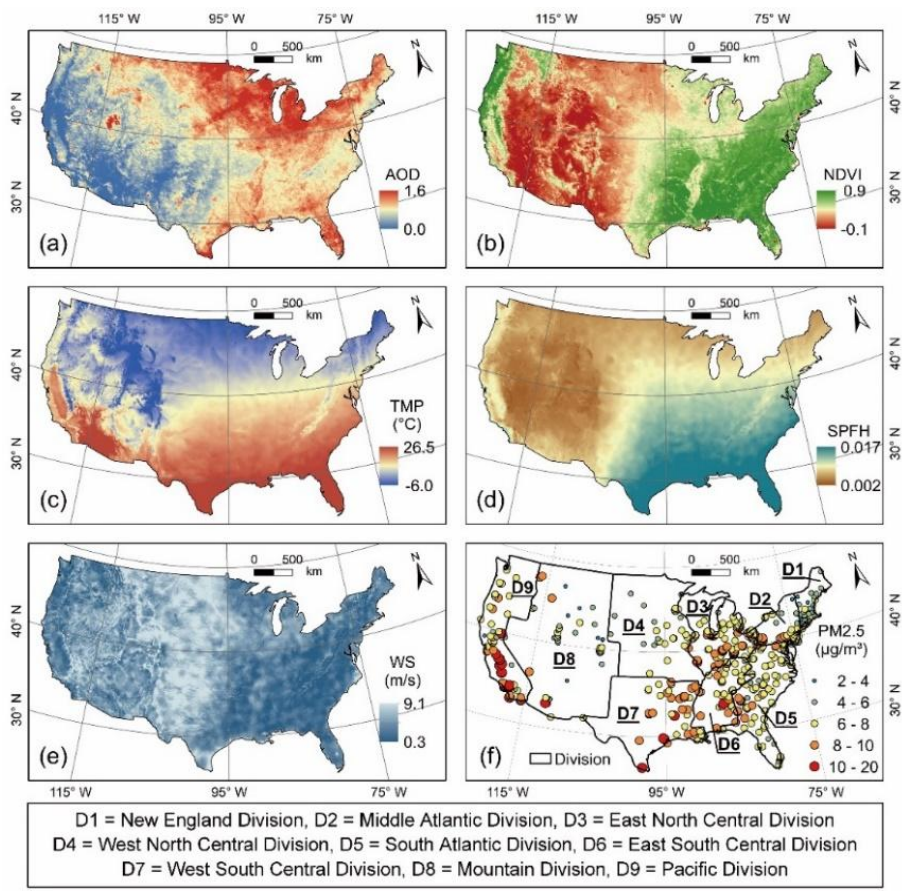

Figure 7. Spatial distribution of PM2.5 concentration and drivers at ground stations over the CONUS in 2019. (The study region can be divided into 9 divisions, namely New England Division (D1), Middle Atlantic Division (D2), East North Central Division (D3), West North Central Division (D4), South Atlantic Division (D5), East South Central Division (D6), West South Central Division (D7), Mountain Division (D8), and Pacific Division (D9))

### 3.2.2 Regression performance of AGWNN

Table 2 presents the regression results of the six models (MLR, ANN, GWR, GWANN, GNNWR, and AGWNN) for the PM2.5 data over the CONUS in the four seasons of 2019. When comparing the regression results of different models vertically, it becomes evident that AGWNN (0.774, 0.702, 0.891, 1.179) outperforms MLR (1.462, 1.385, 2.036, 2.332), ANN (1.325, 1.163, 1.811, 2.232), GWR (0.829, 0.864, 1.096, 1.434), GWANN (0.826, 0.781, 1.079, 1.423), and GNNWR (0.798, 0.732, 0.914, 1.290) in terms of RMSE. Upon comparing the regression results horizontally, the differences among the four seasons are minimal, and the order of goodness-of-fit is as follows: autumn (adjR$^2$=0.871) > summer (adjR$^2$=0.847) > winter (adjR$^2$=0.834) > spring (adjR$^2$=0.799).

Table 2. Regression results of PM2.5 over the CONUS during four seasons of 2019.

| Model | Spring | Summer | Autumn | Winter |
| --- | --- | --- | --- | --- |

|        | RMSE  | adjR² | RMSE  | adjR² | RMSE  | adjR² | RMSE  | adjR² |
|--------|-------|-------|-------|-------|-------|-------|-------|-------|
| MLR    | 1.462 | 0.283 | 1.385 | 0.405 | 2.036 | 0.325 | 2.332 | 0.351 |
| ANN    | 1.325 | 0.411 | 1.163 | 0.580 | 1.811 | 0.466 | 2.232 | 0.406 |
| GWR    | 0.829 | 0.769 | 0.864 | 0.768 | 1.096 | 0.804 | 1.434 | 0.755 |
| GWANN  | 0.826 | 0.771 | 0.781 | 0.811 | 1.079 | 0.810 | 1.423 | 0.758 |
| GNNWR  | 0.798 | 0.786 | 0.732 | 0.834 | 0.914 | 0.864 | 1.290 | 0.801 |
| AGWNN  | 0.774 | 0.799 | 0.702 | 0.847 | 0.891 | 0.871 | 1.179 | 0.834 |

Note: RMSE is in μg/m³.

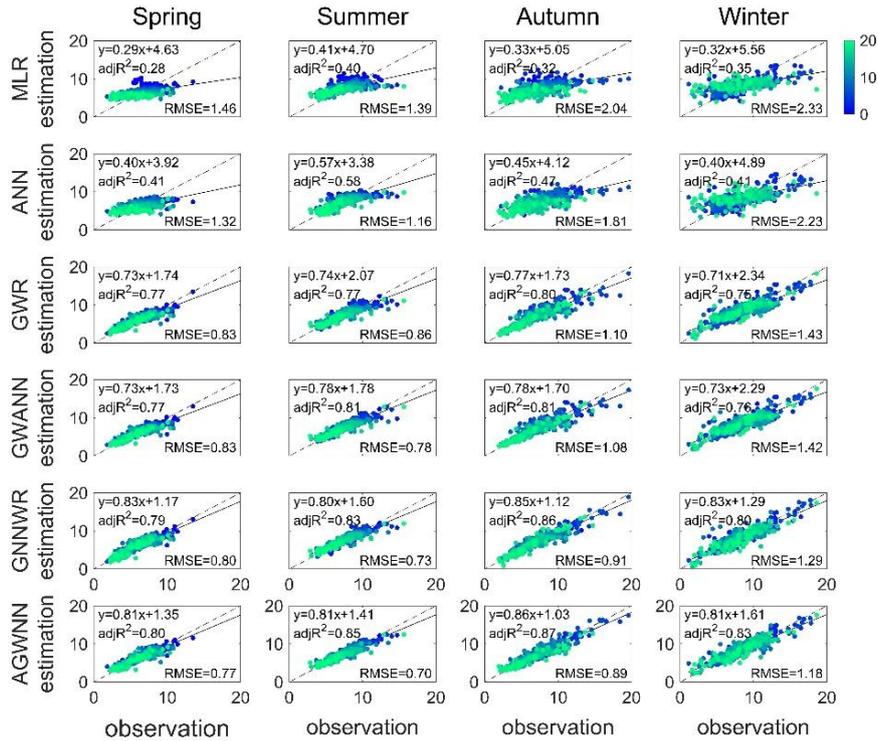

Figure 8. Scatter plots of estimated versus observed PM2.5 for different models.

Figure 8 illustrates the scatter plots of PM2.5 estimation versus observation for the six models, as well as the evaluation metric scores for the best models. The environmental modeling results demonstrate that the fitting performance progressively improves from the MLR to the AGWNN model. The adjR² of GWR ranges between 0.75 and 0.81, indicating a moderate performance relative to all other models. Compared with the traditional GWR method, AGWNN demonstrates very high fitting ability, with adjR² improving by more than 0.02 and RMSE decreasing by more than 0.03. These results indicate that AGWNN achieves excellent generalization performance and accurately reproduces the original state of the PM2.5-AOD model. In conclusion, AGWNN outperforms the other models in all four seasons; thus, AGWNN will be selected as the optimal model for subsequent experiments.

**3.2.3 Spatial heterogeneity of regression coefficients**

The maps (Figure 9) illustrate a comparison of the distribution of regression coefficients for GWR and AGWNN. From the comparative results, it is evident that AGWNN achieves significantly higher regression accuracy than GWR and provides a more detailed and comprehensive reconstruction of fine features. Furthermore, this experiment allows the analysis of the spatial distribution characteristics of regression coefficients from a temporal and spatial perspective. When comparing the seasonal sequences, the spatial heterogeneity characteristics of the same regression coefficient remain largely consistent, demonstrating the strong stability and robustness of AGWNN in reproducing the model. Vertical comparisons allow the spatial heterogeneity distribution among different regression coefficients.

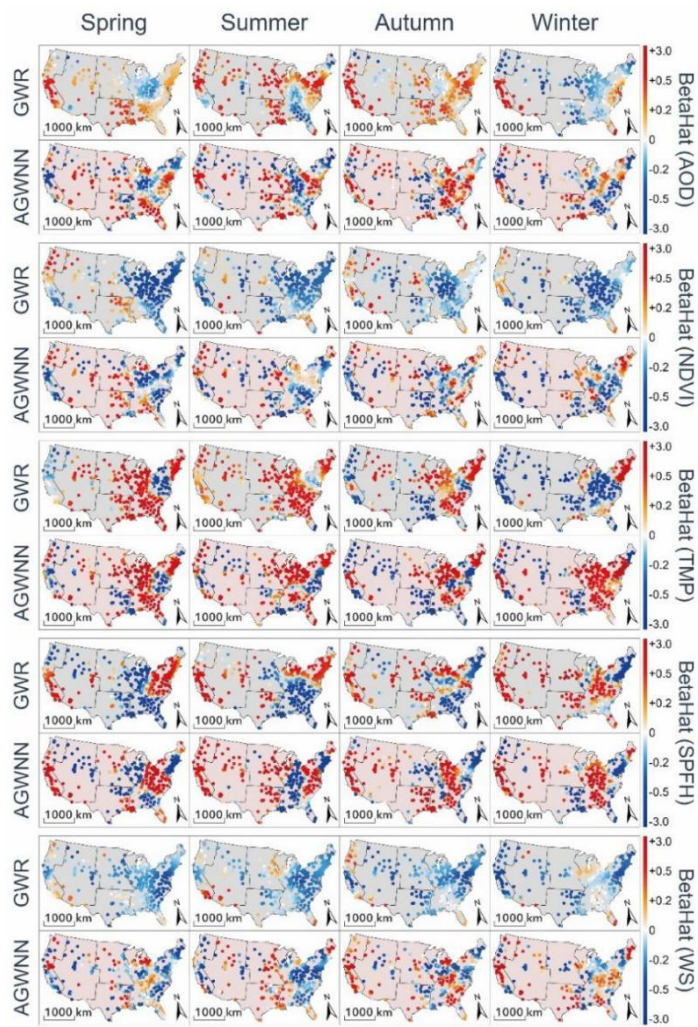

Figure 9. Spatial distribution of regression coefficients in GWR vs. AGWNN models. (Rows 1, 3, 5, and 7 represent the regression coefficients for GWR, while rows 2, 4, 6, and 8 showcase the regression coefficients of AGWNN)

Geographical divisions appearing in this description are replaced by code names, as shown in Figure 7. AGWNN accurately detects the spatial heterogeneity features of the regression coefficients (Figure 9). In autumn, the positive effect of AOD is strengthened, predominantly concentrated at the junction of D6 and D3, most regions of D5, as well as parts of D8 and D9. NDVI exhibits a predominant negative effect during winter, with negative values spanning much of D8, D4, D7, D5, and D9. TMP demonstrates its maximum influence in spring, with a dominant positive effect spanning almost the entirety of the CONUS, except for the upper portions of D7 and D5, and the junction of D8 and D9. During autumn, SPFH experiences a pronounced positive effect, with evident clustering observed in the eastern part of D7 and the middle to lower sections of D9. In summer, WS manifests a notable negative effect, with negative values concentrated at the junction of D6, D5, D3, and D2.

**3.2.4 Estimation diagnosis of PM2.5**

Table 3 shows the comparison results between the AGWNN estimation and the ground-based observation for PM2.5. The AGWNN's estimation (mean=6.633, 7.885, 7.537, 8.133) are close to ground-based observations (mean=6.511, 7.962, 7.557, 8.091), with average PM2.5 fluctuating and increasing from spring to winter. It illustrates the spatial distribution of regression PM2.5 in the CONUS for the four seasons (Figure 10). The spatial distribution of PM2.5 generally exhibits a decreasing trend from coastal to inland areas, with the western and southeastern regions being more susceptible to air pollution compared to the central and northern region. Regions with a Mediterranean climate and tropical desert climate generally exhibit higher PM2.5 concentrations. The western, northeastern, and southern regions constitute the three major industrial zones in the United States and often emit substantial particulate pollutants. When comparing the seasons, the spatial distributions of PM2.5 in the left and right sides of the Figure 10 are basically consistent, indicating that AGWNN has good model reconstruction capability. Overall, air quality in the CONUS is generally stable over the last three seasons. Air quality in the west progressively deteriorated with the flow of the seasons, while air pollutants in the east shifted from the south to the north. The winter of 2019 outbreak of COVID-19 pandemic, a world public health accident, made it difficult for the U.S. to be spared (Berman and Ebisu 2020). The northeastern and western coastal areas are major population and urban agglomerations, yet PM2.5 accumulation is occurring during this period, which undoubtedly increases the health risks to local residents.

Table 3. Comparison between observations and estimations for PM2.5 via AGWNN in 2019.

| Season | Observation (ground station) (µg/m³) | | | Estimation (AGWNN) (µg/m³) | | |
| --- | --- | --- | --- | --- | --- | --- |
| | min | mean | max | min | mean | max |
| Spring | 1.829 | 6.511 | 13.507 | 2.937 | 6.633 | 11.533 |
| Summer | 2.867 | 7.962 | 15.621 | 3.862 | 7.885 | 12.483 |
| Autumn | 1.953 | 7.557 | 19.597 | 2.919 | 7.537 | 17.495 |
| Winter | 1.173 | 8.091 | 22.820 | 2.177 | 8.133 | 17.592 |

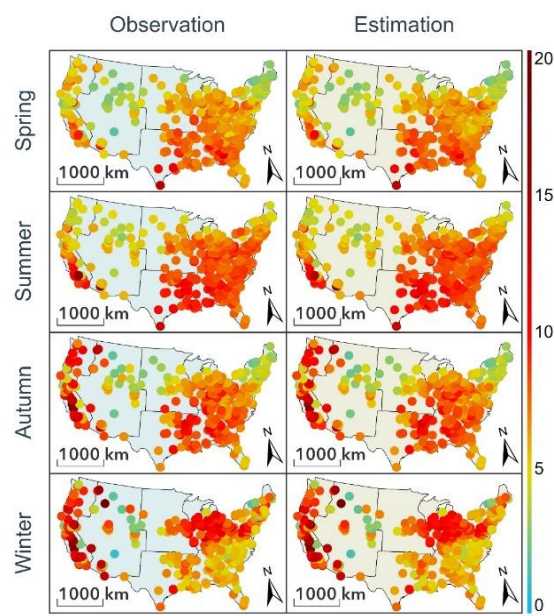

Figure 10. Comparison of ground-based observations and AGWNN estimations for PM2.5 in four seasons.

## 4. Discussion

In this study, we propose a GWL to capture spatial heterogeneity information. Based on this concept, we design AGWNN architecture with input, geographically weighted, hidden, and output layer to learn complex nonlinear relationships. We evaluate AGWNN using synthetic data. When the synthetic data exhibits nonlinearity and high spatial variance, AGWNN has better predictive performance than MLR, ANN and GWR. On one hand, the AGWNN structure addresses global nonlinear regression problems through its traditional hidden layer and output layer. GWL emulates the local geographically weighting calculation process of GWR by mapping neurons to specific spatial locations, equipping neural networks with the capability to discern and analyze spatial heterogeneity.

The GWANN and GNNWR methods leverage the learning capabilities of neural networks to enhance generalization for spatial heterogeneity fitting problems. The AGWNN model is evaluated and compared with these two models using a combination of simulated data and real environmental data in this experiment. The results demonstrate that the AGWNN model outperforms both the GWANN and GNNWR models in terms of fitting and prediction performances, while also exhibiting improved efficiency in model training. In terms of spatial heterogeneity fitting, AGWNN proposes a GWA tuning mechanism, which performs nonlinear optimization on the basis of Gaussian-based geographical weights. In terms of the overall model structure, AGWNN exhibits a high degree of loose coupling. Any combination of GWL and neural networks can give rise to an infinite number of geographically weighted neural networks, enabling adaptation to diverse data scales. With this advanced network structure, AGWNN can be arbitrarily adjusted to avoid overfitting and improve training efficiency. Therefore, the high prediction accuracy and efficiency of AGWNN can be attributed to the generalized expression of spatial heterogeneity and the loosely coupled network structure it possesses.

In addition, it is noteworthy that this study employs both synthetic data and real environmental data to visually represent the spatial heterogeneity relationship of AGWNN. The specific implementation has been elucidated in [Section 2.3.2](#), and this innovative design will contribute novel ideas for the spatial visualization expression of neural networks. Hence, we can confidently assert that AGWNN has been ingeniously designed in terms of network structure and calculation method, effectively preserving the nonlinear fitting capability of ANN as well as GWR's computational ability for spatial heterogeneity relationships.

Evidently, there are still some limitations in the research results that need to be considered in future research or applications. The GWL initialization is based on the geographical weights of existing samples, which can easily lead to structure bloat due to the matching characteristics of GWL neurons and spatial location. In order to address this issue, for a small number of samples, the neurons in GWL correspond to their actual spatial locations, while for a large number of samples, the neurons correspond to virtual spatial locations. This design enhances the flexibility of GWL structure to some extent; however, we acknowledge that complex comparison calculations are necessary to find a more suitable GWL structure. Future plans include optimizing the way geographical weights are obtained and integrating the geographical weight optimization process into AGWNN internally. Additionally, there are plans to expand AGWNN to MGWR and GTWR domains and optimize the network structure to reduce reliance on computational resources. The goal is for the AGWNN model to be applicable in a wider range of scenarios and fields.

**5. Conclusion**

In this study, we propose a GWL in which each neuron is assigned to a specific geographical location for the purpose of computing spatial heterogeneity relationships. A novel and flexible AGWNN, which integrates GWL with ANN structure, is proposed to effectively model spatial heterogeneity and capture nonlinear relationships. The simulation and environmental data demonstrate that AGWNN exhibits superior fitting and prediction capabilities compared to MLR, ANN, GWR, GWANN, and GWNNR in the presence of nonlinearity and spatial variability. The loosely coupled network architecture of AGWNN is applicable to computing scenarios with varying data volumes. Surprisingly, this ingenious structural design of AGWNN also enables it to reliably visualize spatial heterogeneity. In the future, additional experiments across various domains are essential to more thoroughly explore the potential uses of the AGWNN model. In addition, the geographical weight initialization of GWL relies on known samples. The optimization of the model structure and parameters will also be a focus of future work.

## Acknowledgments

This study appreciated anonymous reviewers for their comments and suggestions, which helped improve the quality of our manuscript.

## Disclosure statement

The authors declare that they have no known competing financial interests or personal relationships that could have appeared to influence the work reported in this paper.

## Data and codes availability statement

The data and codes are shared at https://github.com/sd19891020/AGWNN.